\documentclass[letterpaper, 10 pt, conference]{ieeeconf}  

\usepackage{cite}
\usepackage{amsmath,amssymb,amsfonts}
\usepackage{algorithmic}
\usepackage{graphicx}
\usepackage{textcomp}
\usepackage{xcolor}
\usepackage{subcaption}
\usepackage{url}
\usepackage{flafter} 
\usepackage{pifont} 
\newcommand{\cmark}{\ding{51}}
\newcommand{\xmark}{\ding{55}}

\newcommand{\NenvCollision}{16} 
\newcommand{\NrunCollision}{10}   

\usepackage{multirow, makecell, amssymb} 
\usepackage{booktabs} 
\usepackage{siunitx}

\usepackage{hyperref}

\definecolor{mydarkred}{rgb}{0.6,0,0}
\definecolor{mydarkgreen}{rgb}{0,0.6,0}

\hypersetup{
    colorlinks=true,
    linkcolor=mydarkred,
    citecolor=mydarkgreen
}

\newcommand{\fmtmem}[1]{\num[scientific-notation=false,round-mode=places,round-precision=1]{#1}}

\renewcommand{\arraystretch}{1.08}
\IEEEoverridecommandlockouts                              

\title{\LARGE \bf
GPUSimBench: Towards Scalable and Reliable GPU-Accelerated Simulators in Embodied AI
}

\author{Huzhenyu Zhang$^{1,2,*}$, Shenghai Yuan$^{3,*}$,~\IEEEmembership{Member,~IEEE}, Wenrui Yan$^{1}$, \\
Li Ma$^{1}$, Hengjie Li$^{1}$, Jingcheng Pang$^{4,\dagger}$, and Dmitry Yudin$^{2,5,\circ }$%
\thanks{$^{\dagger}$Corresponding author. $^{1}$Shanghai AI Laboratory, China.  $^{2}$MIRAI, Russia. $^{3}$Nanyang Technological University, Singapore. $^{4}$Nanjing University, China. $^{5}$AXXX, Moscow, Russia. Email: $^{*}$zhanghuzhenyu@pjlab.org.cn / $^{*}$shyuan@ntu.edu.sg  / $^{\dagger}$pangjc@lamda.nju.edu.cn / $^{\circ }$yudin.da@miriai.org}%
}

\begin{document}

\maketitle
\thispagestyle{empty}
\pagestyle{empty}

\begin{abstract}
Data-driven embodied AI is rapidly transitioning into a paradigm that scales training through massively parallel simulation, where GPU-accelerated simulators serve as the foundational data infrastructure. However, as computational throughput scales, the underlying trade-offs between parallel efficiency, physical fidelity, and execution determinism remain largely unexamined, hindering the development of reliable robot learning. In this paper, we expose the hidden limits of mainstream GPU-based robotic simulators (e.g., Isaac Lab, Genesis) by introducing GPUSimBench, which focuses on scalability, physical consistency, and computational determinism. First, GPUSimBench establishes a physical grounding evaluation with a controlled inclined-plane task, quantifying the distributional alignment between simulated dynamics and their real-world counterparts. Second, we benchmark parallel scalability by measuring throughput and memory footprints across scaling environment counts. Crucially, beyond standard performance metrics, we unveil and quantify the inherent non-determinism introduced by GPU-batched execution, characterized by significant run-to-run and inter-environment variability even under identical initial conditions. Finally, we identify four empirical regimes of stochasticity within current simulator stacks, highlighting that unbounded scaling can compromise reproducibility without explicit constraints.
\end{abstract}


\begin{figure}[t!]
  \centering
  \includegraphics[width=\linewidth]{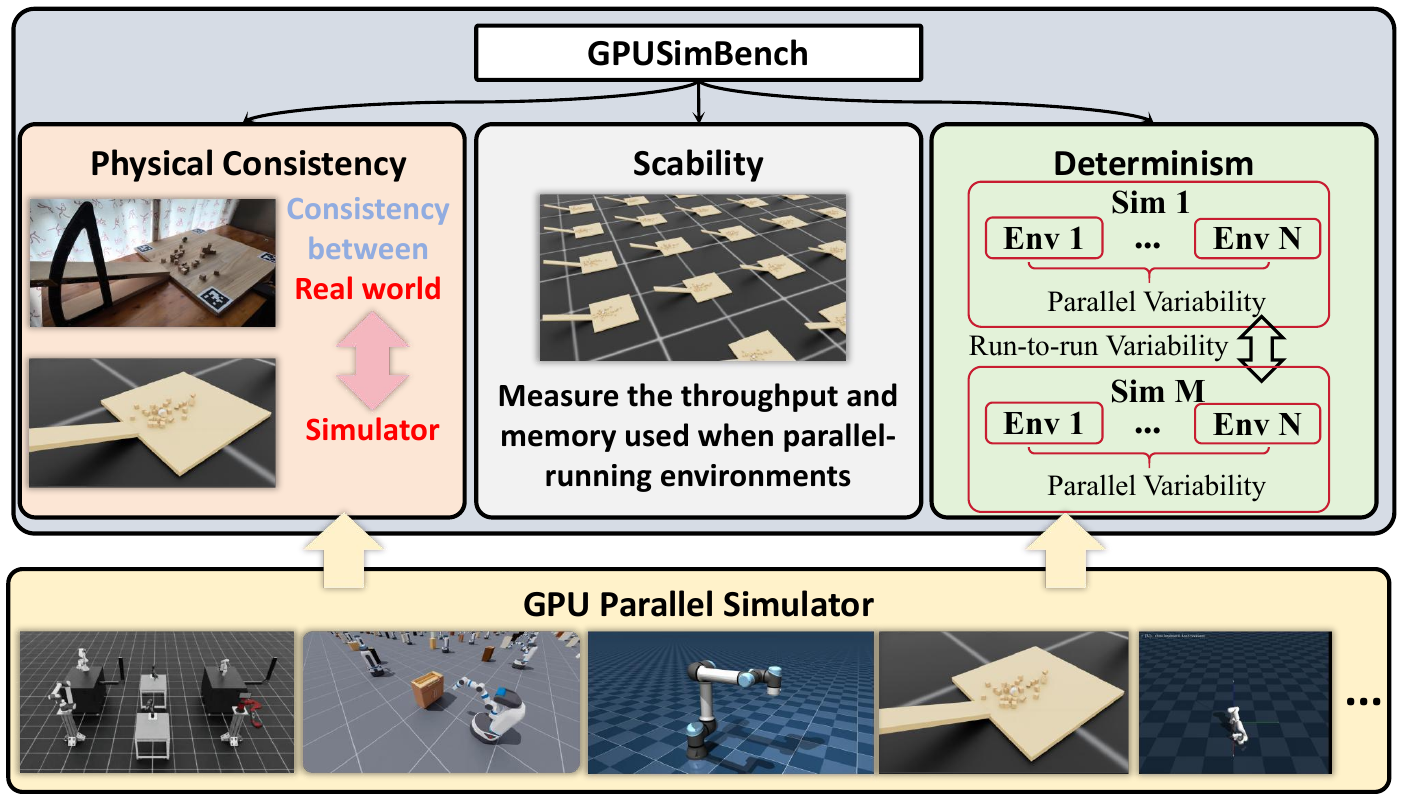}
  \caption{An overview of key features of GPUSimBench: \textit{physical consistency} between simulator and the real world, \textit{scalability} for sampling efficiency, and \textit{determinism} quantified by parallel and run-to-run variability.}
  \label{fig:framework}
\end{figure}

\bstctlcite{IEEEexample:BSTcontrol}

\section{Introduction}

Enabling robots to complete various tasks is a hallmark of machine intelligence. Recent advances in embodied AI take a solid step towards this direction \cite{rt-x,rt-2,rt-1,reviwo,o3f}, which is rapidly transitioning into a paradigm that scales training from extensive robotic data \cite{robot_scaling1,robot_scaling2,robot_scaling3}. In this area, massively parallel simulation has emerged as the foundational data infrastructure. Recent advancements in GPU-accelerated robotic simulators, such as Isaac Lab \cite{makoviychuk2021isaacgym}, and Genesis \cite{Genesis}, enable the simulation of even millions of environments simultaneously. By leveraging the immense parallel computing of modern GPUs, these platforms provide promising computational throughput, fundamentally shifting how the community scales robot learning and policy training \cite{training_parallel1,training_parallel2,training_parallel3,training_parallel4,training_parallel5}.


However, treating these massively parallel simulators merely as infinite data generators overlooks critical underlying complexities. As computational throughput scales to new heights, the complicated trade-offs between parallel efficiency, physical fidelity, and execution determinism remain largely unexamined. In traditional sequential simulators, computational determinism and reliable physical contact modeling are often assumed as baselines. In contrast, the highly concurrent nature of GPU-batched execution introduces subtle numerical variations, parallel synchronization artifacts, and floating-point non-determinism. \textit{Without a rigorous understanding of these phenomena, the development of reliable and reproducible robot learning policies is largely hindered.} Despite the critical role of simulation infrastructure, existing benchmarks primarily focus on task-level performance or parallel capability \cite{geng2025roboverse,sedlacek2025realm,liu2023physicsengines,yoon2023comparative,gpuparallel_benchmark}, such as success rates or convergence speed, rather than the fundamental properties of the simulator engine itself. \textit{There is a notable lack of standardized evaluation for how physical consistency degrades as parallelization increases, or how the GPU-parallelization affects the simulation performance.}

To address these challenges, we introduce GPUSimBench, a comprehensive benchmark designed to expose the hidden limits of mainstream GPU-based robotic simulators. Unlike traditional task benchmarks, GPUSimBench focuses on three key features: scalability, physical consistency, and computational determinism. First, we establish a physical consistency evaluation using a controlled inclined-plane task on both the simulator and the real world, which allows us to quantify the distributional alignment between simulated dynamics and real-world physics. Second, we benchmark parallel scalability by rigorously measuring throughput and memory footprints across varying environment counts. Crucially, beyond standard performance metrics, we unveil and quantify the inherent non-determinism introduced by GPU-batched execution. 
We make a comprehensive evaluation of mainstream GPU-accelerated parallel simulators, including IsaacLab \cite{mittal2025isaaclab}, ManiSkill \cite{taomaniskill3}, Genesis \cite{Genesis}, Madrona \cite{shacklett23madrona}, MuJoCo Warp \cite{warp2022}, MJX \cite{todorov2012mujoco}, and MuJoCo Playground \cite{mujoco_playground_2025}. Simulators that only support CPU-parallel execution or rely on GPUs merely for partial acceleration within the simulation pipeline, such as Gazebo \cite{koenig2004design} and PyBullet \cite{coumans2021}, are not considered in this benchmark.
Our analysis characterizes significant run-to-run and inter-environment variability, even under identical initial conditions.
Specifically, we identify four empirical regimes of stochasticity within current simulator stacks, highlighting how noise accumulates as parallelization increases.
Based on these findings, we provide practical guidelines for simulator selection.

\noindent\textbf{Our main contributions are summarized as follows.}
\begin{itemize}
    \item We expose critical but underexamined limitations of mainstream GPU-based simulators, revealing fundamental trade-offs among throughput, physical fidelity, and execution non-determinism.
    
    \item We introduce \textsc{GPUSimBench}, a unified benchmarking suite for systematically evaluating scalability, sim-to-real physical consistency, and stochasticity in massively parallel robotic simulation.
    
    \item Through controlled cross-simulator and sim-to-real experiments, we quantify GPU-batched non-determinism, identify four empirical stochasticity regimes, and derive practical guidance for simulator choice and reliable robot learning.
\end{itemize}

\section{Overview of Simulation Platforms}

Current GPU-accelerated parallel simulators can be categorized by their underlying programming abstractions and execution models. This classification reflects how simulation state is managed and how parallel work is scheduled onto GPU hardware. Table~\ref{tab:gpu_sim_frameworks} summarizes key design features of the simulators considered in this paper.

\subsection{Array-Based and XLA-Accelerated Simulators}
This category leverages tensor-based programming models, such as JAX \cite{jax2018github}, where the simulation state is represented as large contiguous arrays. Representative systems include \textbf{Brax} \cite{freeman2021brax}, \textbf{MJX}, and \textbf{MuJoCo Playground} \cite{mujoco_playground_2025}. Using JAX-XLA, these simulators compile simulation logic into highly optimized kernels. While this enables seamless integration with deep learning pipelines and automatic differentiation, the static-shape requirements of XLA can necessitate padding for irregular data structures, such as contact buffers, to maintain a fixed memory layout during execution.

\subsection{High-Level Kernel Framework-Based Simulators}
These simulators are built on specialized high-performance computing frameworks that allow users to write simulation logic in a Pythonic syntax while compiling to specialized GPU programs.
\textbf{Genesis} \cite{Genesis} utilizes the Taichi \cite{hu2019taichi}, enabling flexible deployment of physics solvers across backends. Similarly, \textbf{MuJoCoWarp} and \textbf{Taccel} \cite{li2025taccel} are constructed on NVIDIA Warp \cite{warp2022}, a differentiable simulation framework that supports kernel compilation and CUDA Graph capture for low-overhead execution. These platforms balance Python scripting ease with native-level performance, though they require environment authors to explicitly manage data layouts for efficient GPU batching.

\subsection{Task-Centric Simulators}
This category includes platforms that wrap established high-performance physics engines with specialized management layers. \textbf{Isaac Gym} \cite{makoviychuk2021isaacgym} pioneered large-scale GPU-based physics simulation for robot learning, and recent task-centric successors such as \textbf{Isaac Lab} \cite{mittal2025isaaclab} and \textbf{ManiSkill} \cite{taomaniskill3} provide higher-level task APIs while utilizing the GPU-accelerated PhysX backend for rigid and articulated body dynamics. ManiSkill is built on SAPIEN \cite{Xiang_2020_SAPIEN} and supports a wide range of robot embodiments and tasks. These frameworks keep simulation data in device memory to reduce host--device synchronization overhead during parallel execution.

\subsection{ECS-Based Specialized Simulators}
A novel approach is represented by \textbf{Madrona} \cite{shacklett23madrona}, which employs an Entity Component System (ECS) architecture. Unlike the tensor-based approach, Madrona decouples components from systems and schedules simulation work directly on the GPU using a Megakernel. 

\begin{table*}[!t]
\vspace{8pt}
\centering
\caption{Comparison of GPU-accelerated parallel simulation frameworks.
}
\label{tab:gpu_sim_frameworks}
\resizebox{\textwidth}{!}{
\begin{tabular}{lccccccc}
\hline
\noalign{\vskip 2pt}
Feature & Isaac Lab\cite{mittal2025isaaclab} & ManiSkill\cite{taomaniskill3} & Madrona\cite{shacklett23madrona} & Genesis\cite{Genesis} & MuJoCoWarp\cite{warp2022} & MJX\cite{todorov2012mujoco} & Playground\cite{mujoco_playground_2025} \\
\noalign{\vskip 2pt}
\hline
\noalign{\vskip 2pt}
Parallelized Simulation & \cmark & \cmark & \cmark & \cmark & \cmark & \cmark & \cmark \\
Parallelized Heterogeneous Simulation & \cmark & \cmark & \cmark & \cmark & \xmark & \xmark & \xmark \\
Implementation & Isaac Sim & SAPIEN\cite{Xiang_2020_SAPIEN} & Madrona & Genesis & MuJoCo Warp & MJX & Playground \\
Physics Backend & PhysX & PhysX & Custom (XPBD \cite{macklin2016xpbd}) & Taichi-based Custom & Warp \cite{warp2022} & MuJoCo & MuJoCo \\
Primary language & Python & Python & C++ (CUDA Ext.) & Python & Python & Python (JAX) & Python (JAX) \\
\noalign{\vskip 2pt}
\hline
\end{tabular}
}
\end{table*}

\section{Evaluation Methodology}\label{sec:method}
To comprehensively evaluate GPU-accelerated parallel simulators, we design experiments to measure parallel scalability, resource usage, and distribution-level physical fidelity. This section introduces the core metrics used throughout the experiments.

\subsection{Parallel Throughput}
To quantify the computational efficiency of the physics step process, we measure the aggregate simulation throughput in terms of Frames Per Second (FPS) across \(N_{\text{env}}\) parallel environments. In each benchmark, we perform an initial warmup phase that is excluded from timing, followed by a fixed number of physics steps. Let \(N_{\text{step}}\) denote the number of timed steps per environment and \(T_{\text{step}}\) denote the measured time required to execute all calls to steps in \(N_{\text{env}}\) environments. The throughput is defined as
\begin{equation}
    \mathrm{FPS} = \frac{N_{\text{env}} \times N_{\text{step}}}{T_{\text{step}}}.
\end{equation}

\subsection{GPU Memory Usage}
GPU memory usage is a critical indicator of the resource footprint of GPU-accelerated simulators, especially for large-scale parallel runs. Let \(M_{\text{used}}(t)\) denote the instantaneous usage of GPU memory reported by the hardware monitor at time \(t\). We record memory at several checkpoints (baseline before environment creation, post-initialization, post-warmup, peak during stepping, and final). To isolate the memory footprint introduced by simulation, we introduce the peak memory increment
\begin{equation}
    M_{\text{usage}} = M_{\text{peak}} - M_{\text{base}},
\end{equation}
where \(M_{\text{base}}\) is the baseline usage before creating environments and \(M_{\text{peak}}\) is the maximum usage observed. 

\subsection{Distribution Heatmaps and Density Visualization}
We evaluate how accurately different simulators reproduce the final spatial distribution of a cube array after collision by a rolling ball on an inclined plane.
For each simulator, we run \(N_{\text{env}}\) parallel environments with identical initial conditions and record 3D positions of the cube at a fixed physical time \(t_s\).
Given a simulation time step \(\Delta t\), the sampling step is
\begin{equation}
    N_s = \operatorname{round}\!\left(\frac{t_s}{\Delta t}\right).
\end{equation}
Let \(\mathbf{p}_{e,i}\in\mathbb{R}^3\) be the position of the cube \(i\) in the environment \(e\) at \(t_s\).
We aggregate positions across all environments and project them onto the horizontal plane.
\begin{equation}
    \mathcal{P}_{xy} = \bigl\{(x_{e,i},y_{e,i}) \,\big|\, \mathbf{p}_{e,i}=(x_{e,i},y_{e,i},z_{e,i}),\; e=1,\dots,N_{\text{env}}\bigr\}.
\end{equation}
The distribution heat maps are then generated from \(\mathcal{P}_{xy}\) as scatter plots or 2D density maps.

For density visualization, we discretize the \(xy\)-plane into \(K\) fixed-size hexagonal bins.
Let \(c_b\) be the number of points in bin \(b\); we use a logarithmic mapping
\begin{equation}
    D_b = \log_{10}\bigl(\max(c_b, 1)\bigr)
\end{equation}
for color mapping.
This logarithmic density improves high-impact regions and avoids saturation in dense areas, improving comparability for large sample sets.
For comparison of cross-simulators, we compute a global maximum bin count \(c_{\max}\) and use a shared logarithmic color scale over \(1 \leq c \leq c_{\max}\), ensuring consistent color interpretation across all simulators.

\subsection{Distribution Consistency and Real-World Alignment}
We quantify distributional discrepancies using an assignment-based Earth Mover's Distance (EMD) on the projected planar positions. Concretely, we ignore the $z$ direction and compute all distribution distances in $d=2$ on the $xy$ plane.

\subsubsection{Parallel Variability (Intra-run Consistency)}
To assess the stability of a simulator under identical initial conditions, we evaluate Parallel Variability across parallel environments in a single run. For the environment index $e$, let
\begin{equation}
    \mathcal{X}_e = \bigl\{\mathbf{x}_{e,1},\dots,\mathbf{x}_{e,n}\bigr\}, \quad \mathbf{x}_{e,i}\in\mathbb{R}^2,
\end{equation}
denote the planar point set of cube positions at sampling time $t_s$, where $n=m^3$ is the number of cubes per environment ($m=3$ in our setup, hence $n=27$).

The EMD between two environments $e$ and $e'$ is computed via a one-to-one optimal assignment:
\begin{equation}
    W_1(e,e') = \min_{\pi\in S_{n}}\frac{1}{n}\sum_{i=1}^{n} \bigl\lVert \mathbf{x}_{e,i}-\mathbf{x}_{e',\pi(i)}\bigr\rVert_2,
\end{equation}
where $S_{n}$ is the set of all permutations of $\{1,\dots,n\}$. We construct a symmetric consistency matrix $\mathbf{C}\in\mathbb{R}^{N_{\text{env}}\times N_{\text{env}}}$ with elements $\mathbf{C}_{e,e'} = W_1(e,e')$, and summarize it through the average pairwise EMD:
\begin{equation}
    \bar{W}_1^{\text{parallel}} = \frac{2}{N_{\text{env}}(N_{\text{env}}-1)} \sum_{1 \leq e < e' \leq N_{\text{env}}} W_1(e,e').
\end{equation}
This scalar $\bar{W}_1^{\text{parallel}}$ quantifies Parallel Variability: smaller values indicate higher intra-run consistency across parallel environments.

\subsubsection{Real-World Alignment}
A key objective is to measure how closely the simulator output matches the real-world experimental results at the \emph{distribution} level. For a simulator run, we aggregate all parallel environments into a single planar point set $\mathcal{X}^{\text{sim}} = \bigcup_{e=1}^{N_{\text{env}}} \mathcal{X}_e$, and  $\bigl|\mathcal{X}^{\text{sim}}\bigr| = N_{\text{env}}\,n$.

For the real-world apparatus, we perform $K$ independent trials and aggregate all measured cube $xy$ positions into $\mathcal{X}^{\text{real}} = \bigcup_{k=1}^{K} \mathcal{X}^{\text{real}}_{k}$, with $\bigl|\mathcal{X}^{\text{real}}\bigr| = K\,n$. Here we use $N_{\text{env}}=\NenvCollision{}$ and $K=\NenvCollision{}$, so $\bigl|\mathcal{X}^{\text{sim}}\bigr|=\bigl|\mathcal{X}^{\text{real}}\bigr|=N_{\text{env}}\,n$.

The physical-unit EMD between the simulated and real distributions is computed via a one-to-one assignment on the aggregated point sets:
\begin{equation}
    W_1^{\text{phys}}\bigl(\mathcal{X}^{\text{sim}}, \mathcal{X}^{\text{real}}\bigr) = \min_{\pi\in S_{N_{\text{env}}\,n}} \frac{1}{N_{\text{env}}\,n} \sum_{i=1}^{N_{\text{env}}\,n} \Bigl\lVert \mathbf{x}^{\text{sim}}_i - \mathbf{x}^{\text{real}}_{\pi(i)} \Bigr\rVert_2,
\end{equation}
where all positions are expressed in $\mathrm{cm}$ and $\mathbf{x}\in\mathbb{R}^2$.

We define the sim2real gap metric as:
\begin{equation}
    d_{\text{EMD,phys}}^{\text{sim}} = W_1^{\text{phys}}\bigl(\mathcal{X}^{\text{sim}}, \mathcal{X}^{\text{real}}\bigr).
\end{equation}
Smaller $d_{\text{EMD,phys}}^{\text{sim}}$ values indicate better alignment with real-world planar spatial distributions.

\subsubsection{Run-to-Run Variability (Inter-run Consistency)}
To quantify inter-run consistency, let $\mathcal{X}^{(r)}$ be the aggregated planar distribution of run $r\in\{1,\dots,R\}$ (with $R=\NrunCollision{}$), where each $\mathcal{X}^{(r)}$ aggregates all $N_{\text{env}}$ parallel environments of that run. We compute
\begin{equation}
    \bar{W}_1^{\text{run-to-run}} = \frac{2}{R(R-1)} \sum_{1 \leq r < r' \leq R} W_1\bigl(\mathcal{X}^{(r)},\mathcal{X}^{(r')}\bigr),
\end{equation}
where $W_1(\cdot,\cdot)$ uses the same assignment-based EMD as above in $\mathrm{cm}$.

\section{Experiments}
All experiments are conducted on a workstation with a 13th Gen Intel(R) Core(TM) i5-13400F CPU, 32 GB RAM, and an NVIDIA GeForce RTX 5070 GPU with 12 GB memory. The evaluation involves seven physics simulators: ManiSkill (v3.0.0b22), Genesis (v0.3.11), Madrona, MujocoWarp, MJX (MuJoCo v3.4.0), MujocoPlayground (v0.1.0), and IsaacLab (v2.2.1), all retrieved from the main branches of their respective repositories.

\subsection{Parallel Capability}
\begin{figure}[t]
\vspace{8pt}
  \centering
    \includegraphics[width=\columnwidth]{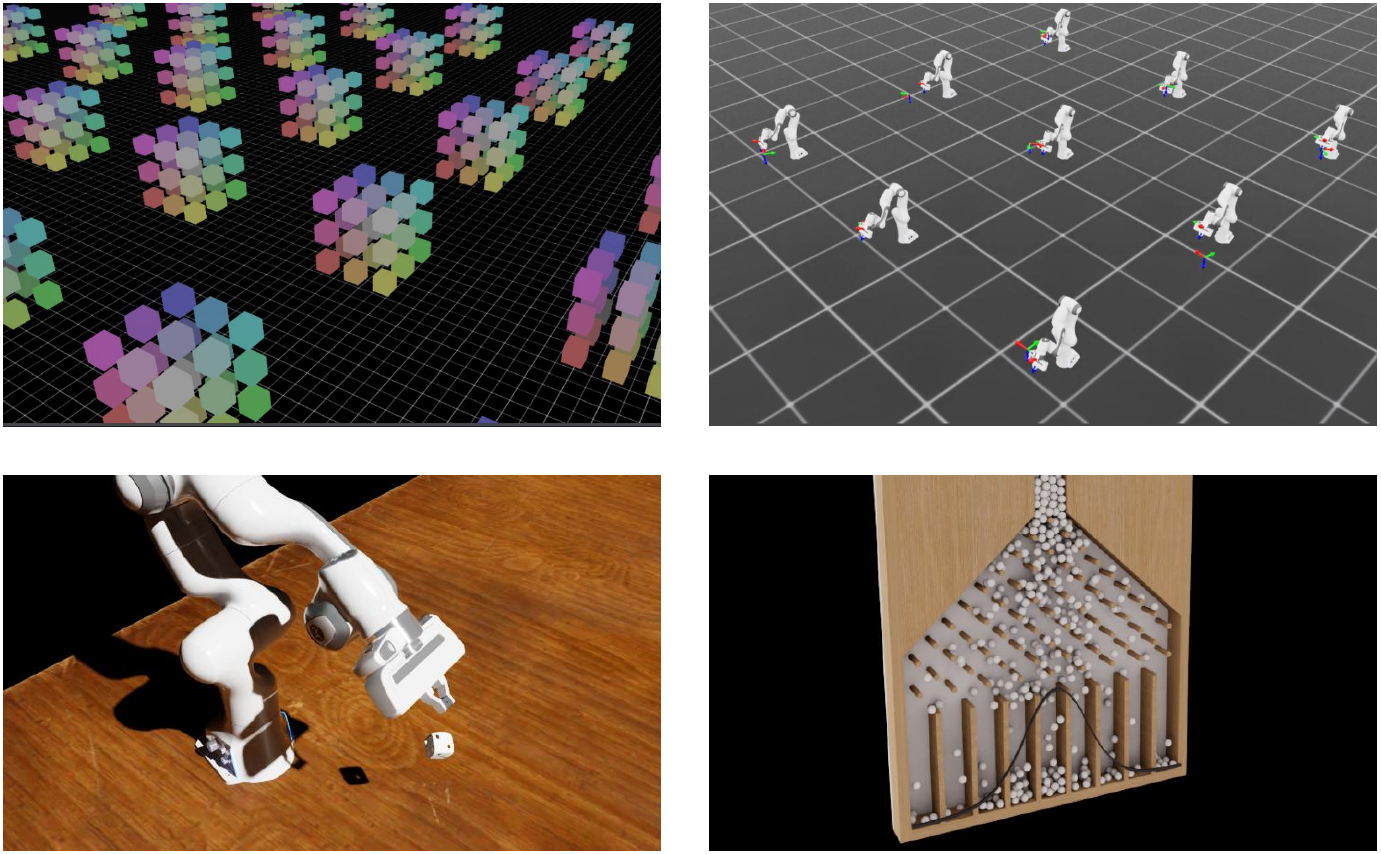}
\caption{Benchmark scenes used to stress parallel physics stepping: (up left) a $3\times3\times3$ cube stack released for free fall and ground impacts, and (up right) a Franka Panda manipulator driven by random actions. Additional real-to-sim distribution experiments, including dice-drop joint pose distribution and Gaussian ball-drop distribution setups, are omitted here due to space constraints and will be released on the project website.}
  \label{fig:exp1_scene_isaaclab}
\end{figure}

We first evaluate the parallel capability of different simulators in two representative scenarios: a free-falling cube array and a Franka manipulator controlled by random actions, as shown in Fig.~\ref{fig:exp1_scene_isaaclab}. For each scenario, we disable rendering and measure \emph{physics stepping time only}. Concretely, we use a unified simulation time step $\Delta t = 0.01$\,s across all simulators and run $100$ warmup steps for JIT compilation or scene initialization. The GPU memory usage of FPS and GPU is measured in $1000$ steps.

\paragraph{Free-falling cube array.} Each environment contains a uniformly stacked cube array with identical geometry and mass properties. The cubes are arranged in a regular grid of dimension \(m \times m \times m\) (with \(m=3\) in our benchmark), a center-to-center spacing equal to the edge length, and a total array height corresponding to \(m\) cubes. The array is initially placed above a flat ground plane and released under gravity to undergo free fall and subsequent impacts with the ground. We use simple cube objects to construct the scene, minimizing the influence of file formats on the parallel-capability measurements.

\paragraph{Franka manipulator with random actions.} Each environment contains a Franka Panda manipulator mounted on a fixed base. The manipulator is initialized in the same configuration and, at each control step, we sample a random \emph{target joint position} uniformly within the physical joint limits as the control signal. And \emph{self-collision} is enabled for all robot links where supported, so that contact interactions are modeled consistently across simulators. Compared with the free-fall scene, this setup introduces a high-DoF articulated mechanism with joint constraints, which stresses articulated-body dynamics, constraint stabilization, and the ability to batch independent control-and-step loops on the GPU.

\subsection{Inclined Collision}


\begin{figure*}[t]
\vspace{8pt}
  \centering
  \includegraphics[width=\textwidth]{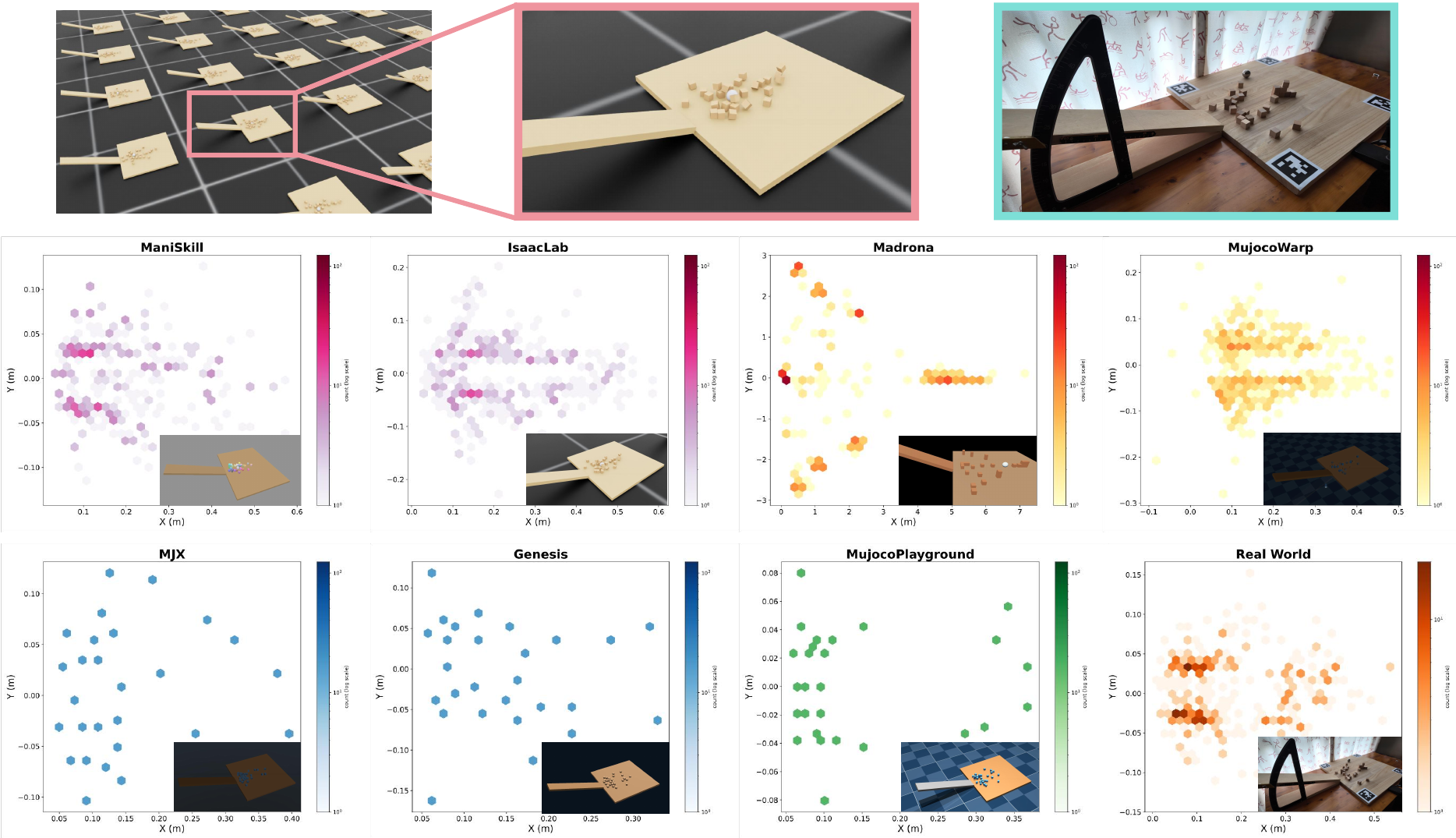}
  \caption{Distribution-level outcomes of the inclined-collision benchmark. For each simulator and the real-world reference, we aggregate final cube positions across $N_{\text{env}}$ parallel environments, project them to the $xy$ plane, and visualize the resulting log-density estimate alongside a representative simulator screenshot. Border colors indicate the determinism regimes Type~1--4 defined in Table~\ref{tab:emd_phys}.}
  \label{fig:density_maps_aggregate}
\end{figure*}

To evaluate physical fidelity and stability under contact-rich dynamics, we design an inclined collision experiment in which the final configuration of a cube array after a ball impact is treated as an observable distribution. Distribution-level outcomes are visualized in Fig.~\ref{fig:density_maps_aggregate} using aggregated density maps and in the right panel of Fig.~\ref{fig:2_27_new} using one-dimensional marginals

\paragraph{Rolling ball down an incline.} The experimental scene consists of a wooden ground plane, an inclined plane, a steel ball, and a cube array. Specifically, the ball rolls down from a specified starting point on the incline, collides with a cube array composed of \(m^3\) uniformly stacked cubes (with \(m=3\)), and slowly stops. We construct a physical experiment platform in the real world and measure key physical parameters, including object sizes, masses, friction coefficients, and restitution as shown in Table~\ref{tab:physical_params}. The real-world experiment is conducted indoors, with the ambient temperature maintained at approximately 15$^\circ$C and the relative humidity maintained at around 15\%.

The material and contact parameters reported below are obtained through identification in professional laboratory facilities and are treated as approximate values. In this benchmark, the real-world distribution is used as an empirical reference of the observed outcome, and the reported EMD values should be interpreted as distribution-level agreement with this measured reference under our best-effort parameter matching. For evaluation, we record all cube positions at a fixed time \(t_s=5.0\,\mathrm{s}\).  

\begin{table}[htbp]
\vspace{8pt}
  \centering
  \caption{Physical Parameters of the Inclined Collision Experiment}
  \label{tab:physical_params}
  \resizebox{\linewidth}{!}{%
    \begin{tabular}{lll}
      \toprule
      Object & Physical Property & Value \\
      \midrule
      \multirow{6}{*}{Ground Plane} 
      & Material & Wood (rough) \\
      & Size & $0.6 \times 0.6 \times 0.02\ \text{m}$ \\
      & Center Coordinates (x,y,z) & $(0.3, 0.0, 0.01)\ \text{m}$ \\
      & Static Friction Coefficient & $0.65$ \\
      & Kinetic Friction Coefficient & $0.45$ \\
      & Coefficient of Restitution & $0.40$ \\
      \midrule
      \multirow{7}{*}{Slope} 
      & Material & Wood (smooth) \\
      & Slope Angle & $20^\circ$ \\
      & Size & $0.6 \times 0.1 \times 0.02\ \text{m}$ \\
      & Center Coordinates (x,y,z) & $(-0.28191, 0.0, 0.12261)\ \text{m}$ \\
      & Static Friction Coefficient & $0.35$ \\
      & Kinetic Friction Coefficient & $0.25$ \\
      & Coefficient of Restitution & $0.40$ \\
      \midrule
      \multirow{7}{*}{Ball} 
      & Material & 304 Stainless Steel \\
      & Radius & $0.0175\ \text{m}$ \\
      & Mass & $0.17782\ \text{kg}$ \\
      & Initial Coordinates (x,y,z) & $(-0.22552, 0.0, 0.13135)\ \text{m}$ \\
      & Static Friction Coefficient & $0.35$ \\
      & Kinetic Friction Coefficient & $0.25$ \\
      & Coefficient of Restitution & $0.55$ \\
      \midrule
      \multirow{6}{*}{Cube} 
      & Material & Wood (smooth) \\
      & Edge Length & $0.02\ \text{m}$ \\
      & Mass (per cube) & $0.005\ \text{kg}$ \\
      & Static Friction Coefficient & $0.35$ \\
      & Kinetic Friction Coefficient & $0.25$ \\
      & Coefficient of Restitution & $0.40$ \\
      \midrule
      \multirow{3}{*}{Cube Array} 
      & Array Dimension (m$\times$m$\times$m) & $3 \times 3 \times 3$ \\
      & Center Spacing (per cube) & $0.02\ \text{m}$ \\
      & Array Center Coordinates (x,y,z) & $(0.08, 0.0, 0.05)\ \text{m}$ \\
      \bottomrule
    \end{tabular}%
  }
\end{table}

\section{Benchmarking Results}

\subsection{Parallel Capability}
The parallel capability benchmark evaluates the simulators' throughput and memory usage across varying environment scales. Two scenarios are considered: a free-falling cube array and a Franka manipulator with random actions. For each simulator, we repeat the experiment for ten independent runs.

\begin{figure*}[!t]
\vspace{8pt}
  \centering
  \includegraphics[width=0.98\textwidth]{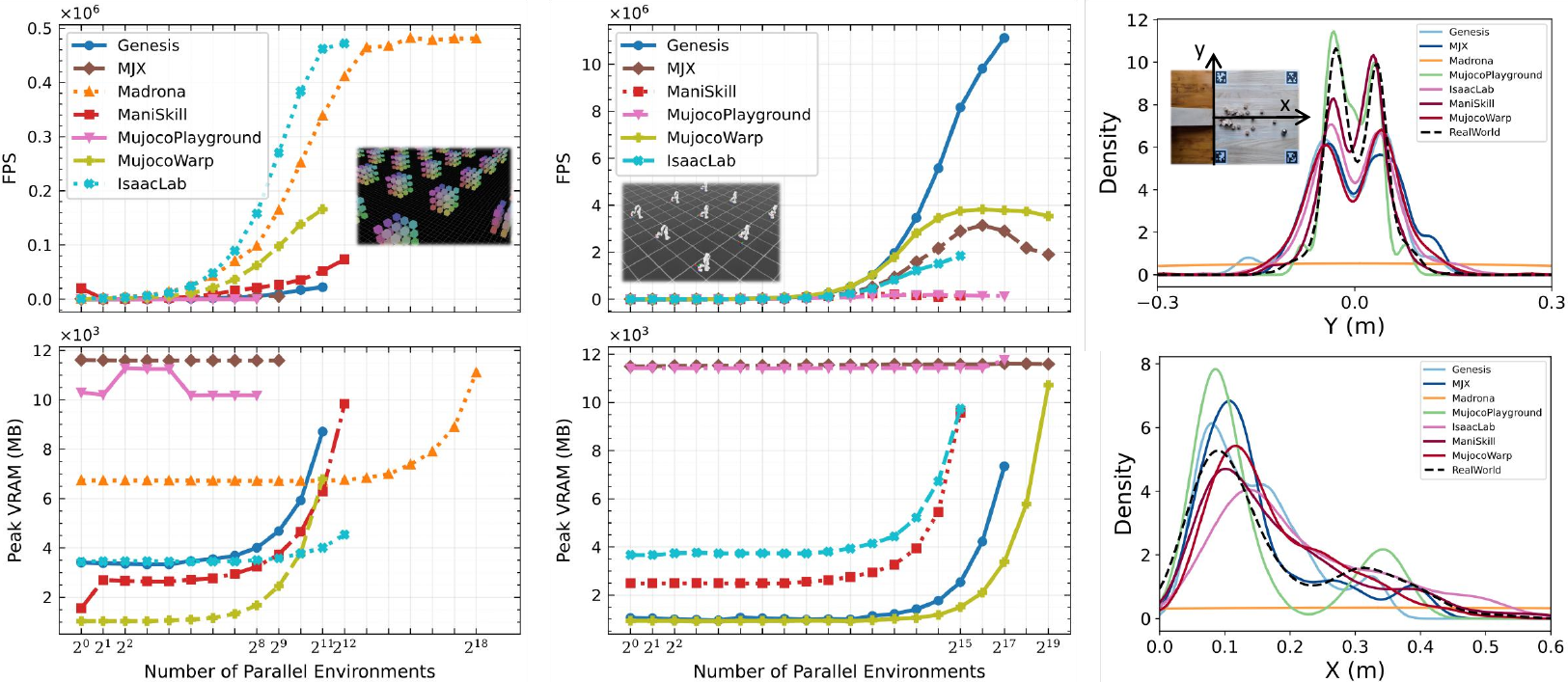}
  \caption{Composite summary of scalability and distribution results. \textbf{Left:} free-fall benchmark throughput and GPU memory increment versus the number of parallel environments $N_{\text{env}}$. \textbf{Middle:} Franka random-action benchmark throughput and GPU memory increment versus $N_{\text{env}}$. Each simulator is evaluated up to its maximum supported parallelism $N_{\text{env}}^{\max}$. \textbf{Right:} one-dimensional marginal distributions along $x$ and $y$ for the inclined-collision benchmark, aggregated across parallel environments for each simulator.}
  \label{fig:2_27_new}
\end{figure*}

\textbf{Throughput \& GPU Memory Usage:}
We summarize the maximum supported parallel environment count $N_{\text{env}}^{\max}$ in each scenario, together with the peak throughput and GPU memory increment observed, in Table~\ref{tab:max_parallelism}. The left and the middle of Fig.~\ref{fig:2_27_new} visualizes the full scalability curves.

\begin{table}[t]
  \centering
  \caption{Maximum supported parallel environment count ($N_{\text{env}}^{\max}$) in the parallel-capability sweep, together with the throughput (FPS) and GPU memory increment (GB) observed at that maximum, for both scenarios. Values are averaged over successful runs.}
  \label{tab:max_parallelism}
  \footnotesize
  \setlength{\tabcolsep}{3.5pt}
  \renewcommand{\arraystretch}{1.05}
  \begin{tabular}{lccc|ccc}
    \toprule
    \multirow{2}{*}{\textbf{Simulator}} & \multicolumn{3}{c|}{\textbf{Free-fall}} & \multicolumn{3}{c}{\textbf{Franka}}\\
    \cmidrule(lr){2-4} \cmidrule(lr){5-7}
    & $N_{\text{env}}^{\max}$ & \makecell{FPS\\($\times 10^{4}$)} & Mem. & $N_{\text{env}}^{\max}$ & \makecell{FPS\\($\times 10^{6}$)} & Mem.\\
    \midrule
    Genesis      & $2^{11}$ & \num[scientific-notation=false,round-mode=places,round-precision=3]{2.233712}  & \fmtmem{7.75}  & $2^{17}$ & \num[scientific-notation=false,round-mode=places,round-precision=3]{11.123920} & \fmtmem{6.42} \\
    Isaac Lab    & $2^{12}$ & \num[scientific-notation=false,round-mode=places,round-precision=3]{47.189328} & \fmtmem{3.34}  & $2^{15}$ & \num[scientific-notation=false,round-mode=places,round-precision=3]{1.851086}  & \fmtmem{8.13} \\
    Madrona      & $2^{18}$ & \num[scientific-notation=false,round-mode=places,round-precision=3]{48.107876} & \fmtmem{9.53}  & \multicolumn{3}{c}{\textit{Not supported}} \\
    ManiSkill    & $2^{12}$ & \num[scientific-notation=false,round-mode=places,round-precision=3]{7.374857}  & \fmtmem{8.34}  & $2^{15}$ & \num[scientific-notation=false,round-mode=places,round-precision=3]{0.149343}  & \fmtmem{8.26} \\
    MJX          & $2^{9}$  & \num[scientific-notation=false,round-mode=places,round-precision=3]{0.534038}  & \fmtmem{10.61} & $2^{19}$ & \num[scientific-notation=false,round-mode=places,round-precision=3]{1.905057}  & \fmtmem{10.63} \\
    MuJoCo Warp  & $2^{11}$ & \num[scientific-notation=false,round-mode=places,round-precision=3]{16.600347} & \fmtmem{5.87}  & $2^{19}$ & \num[scientific-notation=false,round-mode=places,round-precision=3]{3.540859}  & \fmtmem{9.83} \\
    Playground   & $2^{8}$  & \num[scientific-notation=false,round-mode=places,round-precision=3]{0.003472}  & \fmtmem{8.89}  & $2^{17}$ & \num[scientific-notation=false,round-mode=places,round-precision=3]{0.135495}  & \fmtmem{10.28} \\
    \bottomrule
  \end{tabular}
  \renewcommand{\arraystretch}{1.0}
  \normalsize
\end{table}

\subsection{Physical Consistency}

\vspace{-4pt}

\begin{table*}[!t]
\vspace{8pt}
\centering
\caption{Inclined-collision benchmark summary: physical alignment to the real-world apparatus ($d_{\text{EMD,phys}}^{\text{sim}}$), intra-run consistency ($\bar{W}_1^{\text{parallel}}$), and inter-run consistency ($\bar{W}_1^{\text{run-to-run}}$). Checkmarks indicate whether the corresponding variability is present (nonzero after rounding) and are used to define Type~1--4. Values are mean $\pm$ standard deviation over \NrunCollision{} runs (for $d_{\text{EMD,phys}}^{\text{sim}}$ and $\bar{W}_1^{\text{parallel}}$) or over all run pairs (for $\bar{W}_1^{\text{run-to-run}}$).}
\label{tab:emd_phys}\label{tab:parallel_run-to-run}
\footnotesize
\renewcommand{\arraystretch}{1.08}
\setlength{\tabcolsep}{8pt}
\begin{tabular}{c l c c c c c}
\toprule
Type & Simulator & $d_{\text{EMD,phys}}^{\text{sim}}$ (cm) & $\bar{W}_1^{\text{parallel}}$ (cm) & \makecell{Parallel Var.} & $\bar{W}_1^{\text{run-to-run}}$ (cm) & \makecell{Run-to-Run Var.}\\
\midrule
\multirow{2}{*}{1} & Isaac Lab & $5.400\pm0.000$ & $4.21\pm0.00$ & \cmark & $0.00\pm0.00$ & \xmark\\
& ManiSkill & $2.520\pm0.000$ & $4.76\pm0.00$ & \cmark & $0.00\pm0.00$ & \xmark\\
\midrule
\multirow{2}{*}{2} & Genesis & $4.780\pm0.000$ & $0.00\pm0.00$ & \xmark & $0.00\pm0.00$ & \xmark\\
& MJX & $3.970\pm0.000$ & $0.00\pm0.00$ & \xmark & $0.00\pm0.00$ & \xmark\\
\midrule
\multirow{2}{*}{3} & Madrona & $211.60\pm2.22$ & $32.53\pm1.86$ & \cmark & $11.47\pm1.40$ & \cmark\\
& MuJoCo Warp & $4.230\pm0.740$ & $4.55\pm1.39$ & \cmark & $2.15\pm0.82$ & \cmark\\
\midrule
4 & MuJoCo Playground & $3.380\pm0.110$ & $0.00\pm0.00$ & \xmark & $1.39\pm0.47$ & \cmark\\
\bottomrule
\end{tabular}
\normalsize
\end{table*}

The second part of our evaluation focuses on parallel and physical consistency through an inclined collision experiment. The simulated distribution of cubes is compared against real-world distribution using the planar ($xy$) assignment-based Earth Mover's Distance (EMD), i.e., we ignore the $z$ direction. We use $N_{\text{env}}=\NenvCollision{}$ parallel environments per simulator run and perform $K=\NenvCollision{}$ matched real-world trials, so both the simulated and real distributions contain $N_{\text{env}}\,n$ cube $xy$ positions and can be compared with a one-to-one assignment.

Table~\ref{tab:emd_phys} summarizes both physical alignment to the real-world apparatus and consistency metrics in the inclined-collision experiment. In terms of fidelity, ManiSkill and MJX achieve the lowest EMD values of $2.520\pm0.000$\,cm and $3.970\pm0.000$\,cm, respectively. MuJoCo Playground also shows strong accuracy with an EMD of $3.380\pm0.110$\,cm, and Genesis follows closely at $4.780\pm0.000$\,cm. In contrast, Isaac Lab ($5.400\pm0.000$\,cm) and Madrona ($211.60\pm2.22$\,cm) exhibit larger discrepancies. The extreme value of Madrona is caused by its current implementation of the XPBD \cite{macklin2016xpbd} solver in our setup: the cubes do not reliably decelerate in the wooden plane due to insufficient frictional effects, leading to large planar drift. MuJoCo Warp shows a moderate deviation ($4.230\pm0.740$\,cm). Table~\ref{tab:emd_phys} is obtained as mean $\pm$ standard deviation in ten independent runs.

\section{Discussion}

\subsection{Parallel and Run-to-Run Variability}

We use two consistency metrics defined in Section~\ref{sec:method}: (i) \emph{Parallel Variability}, the mean pairwise EMD \(\bar{W}_1^{\text{parallel}}\) between environments within a \emph{single} run, averaged over ten independent runs, and (ii) \emph{Run-to-Run Variability}, the mean pairwise EMD \(\bar{W}_1^{\text{run-to-run}}\) between aggregated distributions from all pairs of independent runs under the same nominal configuration. In Table~\ref{tab:parallel_run-to-run}, the values are reported as the mean $\pm$ standard deviation for ten runs (for Parallel Variability) or for all pairs of runs (for Run-to-Run Variability). Values displayed as $0.00\pm0.00$ indicate the absence of both Parallel Variability and Run-to-Run Variability.

For discussion purposes, we group the simulators into four empirical regimes based on whether each variability metric is present or absent in Table~\ref{tab:parallel_run-to-run}. In all inclined-collision runs, we fix random seeds and disable all task-level randomization, so any observed variability reflects numerical and execution-level effects rather than intentional stochasticity.

\noindent\textbf{Type 1 (Isaac Lab, ManiSkill): Parallel Variability present; Run-to-Run Variability absent.}
Within a single run, different parallel environments drift to measurably different outcome distributions, leading to non-zero Parallel Variability. However, when we repeat the entire experiment across independent runs under the same nominal configuration, the Run-to-Run Variability is reported as $0.00\pm0.00$ in Table~\ref{tab:parallel_run-to-run}. At our reporting precision, the aggregate outcome distribution is effectively unchanged between runs.

\noindent\textbf{Type 2 (Genesis, MJX): Parallel Variability absent; Run-to-Run Variability absent.}
Both metrics are reported as $0.00\pm0.00$ in Table~\ref{tab:parallel_run-to-run}. In our benchmark, this means that the results are indistinguishable between parallel environments and between independent runs at the reported precision, making this regime the most straightforward choice for fair comparisons.

\noindent\textbf{Type 3 (Madrona, MuJoCo Warp): Parallel Variability present; Run-to-Run Variability present.}
Both intra-run (between parallel environments) and inter-run (between runs) variability are measurable in Table~\ref{tab:parallel_run-to-run}. In other words, repeating the same nominal setup can change the aggregate outcome distribution, and environments within a run also diverge from one another.

\noindent\textbf{Type 4 (MuJoCo Playground): Parallel Variability absent; Run-to-Run Variability present.}
Parallel environments within a run are reported as $0.00\pm0.00$, but repeating the experiment across independent runs yields measurable Run-to-Run Variability. 
\begin{table}[!t]
\vspace{8pt}
  \centering
  \caption{Guidelines for simulator choice.}
  \label{tab:simulator_guidelines}
  \renewcommand{\arraystretch}{1.10}
  \setlength{\tabcolsep}{3pt}
  \scriptsize
  \begin{tabular*}{\columnwidth}{@{\extracolsep{\fill}} p{0.70\columnwidth} p{0.26\columnwidth}}
    \toprule
    \textbf{Keywords (task)} & \textbf{Recommended} \\
    \midrule
    Overall performance \& Balanced trade-off & \textbf{Isaac Lab} \\
    Reproducibility \& Fair comparisons \& Repeated runs & \textbf{MJX}, \textbf{Genesis} \\
    Sim-to-real \& Contact-rich \& Repeated-run tuning/reporting & \textbf{MJX} \\
    Sim-to-real \& Contact-rich \& Single-run evaluation & \textbf{ManiSkill}, \textbf{MuJoCo Playground} \\
    Large parallelism \& Articulated robot & \textbf{Genesis} \\
    Large parallelism \& Simple agent & \textbf{Madrona}  \\
    Memory-limited \& Modest parallelism & \textbf{MuJoCo Warp}  \\
    \bottomrule
  \end{tabular*}
  \normalsize
\end{table}

\subsection{Mechanisms Behind the Stochasticity}
In this benchmark, we fix random seeds and disable task-level randomization. Consequently, the observed variability in Table~\ref{tab:parallel_run-to-run} isolates the \emph{numerical and execution-level artifacts} inherent to batched GPU simulation. We suggest that these variability metrics arise from a tightly coupled interplay between hardware-level execution models (SIMT dynamic scheduling), algorithmic solver structures, and the inherently non-smooth nature of rigid-body contact dynamics. We interpret these mechanisms as follows:

\paragraph{Parallel Variability}
Parallel Variability is quantified by $\bar{W}_1^{\text{parallel}}$, with $\bar{W}_1^{\text{parallel}}>0$ indicating divergence between parallel environments within \emph{a single run} under identical initialization. The non-zero $\bar{W}_1^{\text{parallel}}$ is typically introduced by GPU-parallel computations that rely on atomics or parallel reductions. Because floating-point accumulation depends on operation order, different thread schedules can create small numerical noise across environments; in contact-rich rollouts, this noise accumulates and appears as a distribution shift. For instance, the Parallel Variability in Isaac Lab and ManiSkill arises from PhysX’s GPU execution-order differences in floating-point computations, where small round-off perturbations accumulate over contact-rich rollouts.

\paragraph{Run-to-Run Variability}
Run-to-Run Variability is quantified by $\bar{W}_1^{\text{run-to-run}}$, with $\bar{W}_1^{\text{run-to-run}}>0$ indicating differences between \emph{multiple independent runs} under the same nominal configuration. In our setting, a primary source is the algorithmic path taken by iterative numerical solvers: small floating-point perturbations can change contact ordering, activation of constraints, projection decisions, and early-termination conditions, leading to different sequences of intermediate iterates even with identical nominal settings. In long-horizon contact-rich rollouts, these solver-path differences can compound into observable distribution shifts.


\subsection{Guidelines for Simulator Choice}

\begin{figure}[!t]
  \centering
  \includegraphics[width=\columnwidth]{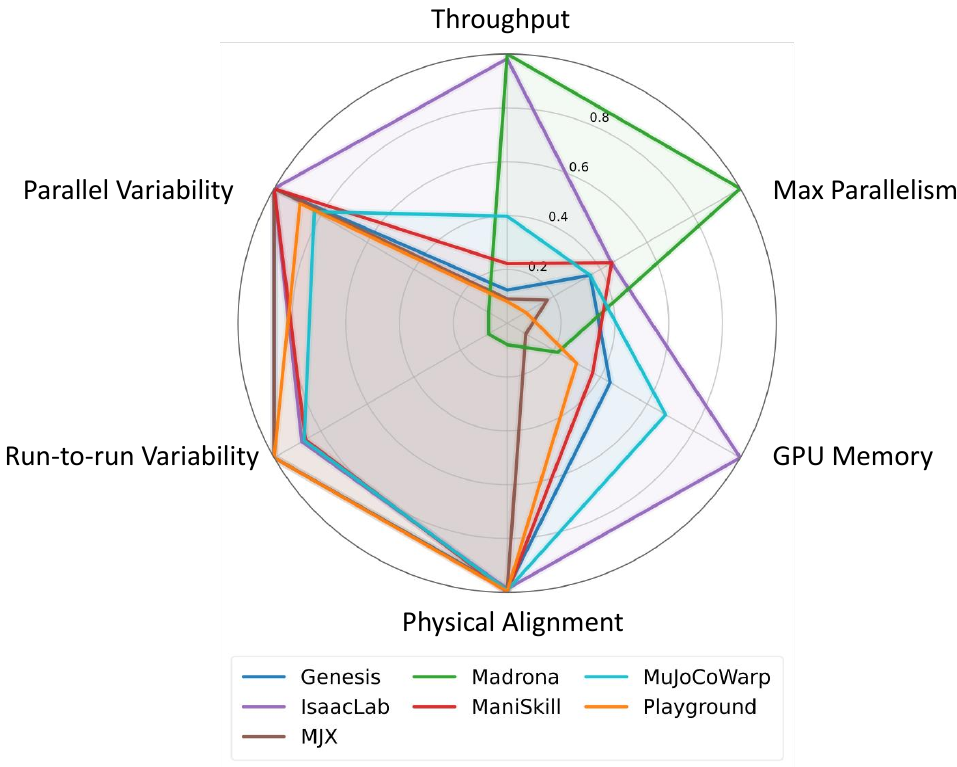}
  \caption{An overall statistics of the key features for different simulators. Larger values indicate better performance.}
  \label{fig:rader}
  \vspace{-10pt}
\end{figure}

Based on scalability and real-world-matched inclined-collision results, Fig.~\ref{fig:rader} compares simulators across six axes: throughput, maximum parallelism, GPU memory footprint, physical alignment (EMD), Parallel Variability, and Run-to-run Variability.
All axes are min--max normalized (with a $\log_2$ transform on maximum parallelism before normalization). For metrics where lower is better (GPU memory footprint, EMD, Parallel Variability, Run-to-run Variability), we invert the direction so that larger radial values always indicate better performance.

We provide the guidelines below based on the variability regimes and the physical-alignment ranking in Table~\ref{tab:emd_phys}. Use Fig.~\ref{fig:rader} to compare shortlisted candidates across these criteria, rather than relying on any single metric.

\paragraph{Sim-to-real transfer and contact-rich manipulation.}
Start by shortlisting simulators with the strongest nominal physical alignment in Table~\ref{tab:emd_phys}. If your workflow involves contact-parameter tuning, design comparisons, or reporting averaged sim-to-real outcomes across repeated runs, additionally prioritize low Run-to-Run Variability in Table~\ref{tab:parallel_run-to-run} to reduce confounding from solver path drift; in our benchmark, MJX offers the most favorable overall trade-off between alignment and run-to-run stability. If you evaluate within a single run and mainly need plausible contact behavior without extensive tuning, ManiSkill and MuJoCo Playground are also competitive choices. Domain randomization can further improve robustness, but it is best applied around a well-aligned nominal model rather than used to compensate for systematic simulator bias.

\paragraph{Scaling-first training with very large parallelism.}
When sample throughput is the primary constraint, choose simulators that maximize parallelism for your target agent class. For multi-joint articulated robots such as Franka, Genesis achieves the highest throughput in our Franka scalability experiment in Fig.~\ref{fig:2_27_new}, making it a strong default for scaling training with articulated agents. For tasks that do not rely on articulated robots or custom agent pipelines, Madrona is a compelling option for pushing maximum parallelism, as its ECS architecture is designed for highly batched GPU execution.

\paragraph{Hardware-limited setups with modest parallelism.}
If GPU memory is the primary bottleneck and only a small number of parallel environments is needed, MuJoCo Warp is a practical choice. In our scalability experiment, it has the lowest GPU memory footprint at low environment counts while maintaining moderate throughput, as shown in Fig.~\ref{fig:2_27_new}.

\paragraph{Balanced default choice.}
When your task does not impose a dominant constraint and prefer a well-balanced simulator, Isaac Lab is a strong default because it offers a competitive balance across throughput, scalability, memory footprint, physical alignment, and reproducibility.

\section{Conclusion and Limitation}
This paper benchmarks mainstream GPU-accelerated parallel robotic simulators under unified conditions, focusing on the practical trade-offs between throughput, GPU memory footprint, and physical consistency. Using two scalability tasks (free-falling cube arrays and a randomly actuated Franka arm) and a real-world-matched inclined-collision experiment, we quantify both performance scaling and distribution-level physical fidelity.
Across simulators, we observe distinct regimes of determinism characterized by Parallel Variability and Run-to-Run Variability, showing that GPU-parallel design choices affect not only speed but also reproducibility and outcome distributions. We release this benchmark and evaluation protocol to support more principled simulator selection and future work on scalable, physically reliable GPU simulation.

This work has several limitations. First, GPUSimBench currently covers only a limited set of tasks and metrics, and the conclusions should be interpreted within this benchmark scope. Specifically, we evaluate two scalability scenes and one contact-rich real-world-matched experiment, all on a single hardware and software stack. As a result, the reported performance and variability may change across different GPUs, drivers, and simulator versions. In addition, the current benchmark does not yet cover deformable-body dynamics, fluid simulation, or the effects of perception and sensor rendering. Extending the benchmark to these settings is an important direction for future work.




\bibliographystyle{IEEEtran}
\bibliography{ref}

\end{document}